\documentclass[10pt, conference, compsocconf]{IEEEtran}
\ifCLASSINFOpdf
\else
\fi
\usepackage{float}
\usepackage{amsthm,subfigure}

\usepackage{subfigure}
\usepackage{adjustbox}
\usepackage{multirow}

\usepackage{hyperref}

\usepackage{graphicx}

\usepackage{amsthm,subfigure}
\usepackage{algorithmic}
\usepackage{ textcomp }

\begin{document}
%
\title{Detecting Events in Crowds Through Changes in Geometrical Dimensions of Pedestrians}


\author{
\IEEEauthorblockN{Matheus Schreiner Homrich da Silva, Paulo Brossard de Souza Pinto Neto, \\ Rodolfo Migon Favaretto, Soraia Raupp Musse}
\IEEEauthorblockA{School of Technology, PUCRS,  \\
Porto Alegre, Rio Grande do Sul \\
Email: {matheus.schreiner,paulo.brossard,rodolfo.favaretto}@acad.pucrs.br,  \\{soraia.musse}@pucrs.br}
}


%




\maketitle

\begin{abstract}
teste
Security is an important topic in our contemporary world, and the ability to automate the detection of any events of interest that can take place in a crowd is of great interest to a population. We hypothesize that the detection of events in videos is correlated with significant changes in pedestrian behaviors. In this paper, we examine three different scenarios of crowd behavior, containing both the cases where an event triggers a change in the behavior of the crowd and two video sequences where the crowd and its motion remain mostly unchanged. With both the videos and the tracking of the individual pedestrians (performed in a pre-processed phase), we use Geomind, a software we developed to extract significant data about the scene, in particular, the geometrical features, personalities, and emotions of each person. We then examine the output, seeking a significant change in the way each person acts as a function of the time, that could be used as a basis to identify events or to model realistic crowd actions. When applied to the games area, our method can use the detected events to find some sort of pattern to be then used in agent simulation. Results indicate that our hypothesis seems valid in the sense that the visually observed events could be automatically detected using GeoMind.

\end{abstract}


%
\IEEEpeerreviewmaketitle

\section{Introduction}
\label{sec:intro}

As the world keeps on showing signs of nonstop violence, security is still a big issue and also an area with many challenges and problems to be solved. With that said, our work primarily focuses on overall safety and  secondarily on how can our work improve game experiences. One important topic in this area is event detection. Many methods have been proposed to deal with that as discussed in Section~\ref{sec:related}. However, in many of them, an event is characterized by a difference in terms of observed pattern, that can be local but in most cases is global~\cite{tran2013video}. In our work, we want to use the individual differences over time in order to detect events, so no training phases of thresholds have to be specified once we observe and extrapolate many aspects of the individuals, such as emotions, personalities and etc.

Our main hypothesis is that GeoMind\footnote{GeoMind can be obtained at \url{https://www.rmfavaretto.pro.br/geomind}} software~\cite{geomind_paper} can be used to detect events in video sequences without the need of training or events specification. GeoMind was developed by Favaretto and aims to extract features from pedestrians in video sequences such as geometric dimensions, personalities, cultural aspects, and emotions status frame-by-frame. As we are going to show in obtained results, GeoMind seems to be an appropriate software and methodology to detect pedestrian events in the video.

The remainder of this paper is organized as follows: Section~\ref{sec:related} briefly describes some methods existent in the literature. Section\ref{sec:method} details the method we used in this paper while Section~\ref{sec:results} presents three short sequences used to discuss our hypothesis.


\section{Related Works}
\label{sec:related}


The understanding of videos containing pedestrians is important in simulation and computer vision areas~\cite{Villamil2003,Paiva2005,Favaretto2016}.
 Regarding unusual event detection in crowds, Adam et al.~\cite{adam2008robust} proposes an approach based on multiple local monitors which collect low-level statistics, and do not rely on object tracking.
Almeida et al.~\cite{de2016detection} uses an approach based on 2D motion histograms in time, and also local effects, to identify motion pattern changes in human crowds that could be related to unusual events in crowds.

Tran et al.~\cite{tran2013video} propose a search for spatiotemporal paths in video space to detect anomalies in video sequences, which is used to detect a range of events, including anomaly event detection, walking person detection, and running detection.  
Mehran et al.~\cite{mehran2009abnormal} proposes a Social Force model to detect unusual crowd behavior; in brief, by means of the average optical flow of the scene, a force flow is calculated, which is used to model the normal behavior of the scene. Using that flow as its reference point, it is then possible to detect and localize abnormal events in the video.

GeoMind~\cite{geomind_paper} is a software capable of extracting emotional, personality, and cultural aspects from a solely thorough the movement of pedestrians; as specified in Section~\ref{sec:method}. It is the tool we have chosen to extract crowd data in this paper.


\section{Method}
\label{sec:method}

This section describes our approach. It is organized in three steps: \textit{i)} Video selection, \textit{ii)} Pedestrian tracking  and \textit{iii)} Output Data.

\subsection{Video Selection}

We searched for videos that started with a crowd behaving normally (structured behavior) at first and ended with the same crowd exhibiting some kind of unusual behavior (e.g. sudden dispersion, erratic movement, etc.) caused by some event. The videos also must be shot from top-of-view vision, with a camera that stays stationary during the course of the video, as a prerequisite of our data extraction software.


\subsection{Pedestrian tracking}

The tracking of the individual pedestrians is not supposed to be a contribution of this work, so we use CVAT tool~\cite{CVAT}, which is a tool to assist in the annotation of video sequences, capable of interpolating the position of pedestrians through keyframes, creating a smooth and accurate representation of motion in image space.
As for the perspective correction, we used the built-in tools from the OpenCV library~\cite{opencv_library}.
Once all extracted information is mapped, we proceed with data analysis, as described in the next section.

\subsection{Analyzed Data}

The output files generated by GeoMind contain all the information about each of the pedestrians present in the video at each frame. These files contain a series of features, such as cultural aspects, personality traits, social aspects, emotions, socialization, isolation, and collectivity, among other characteristics. In our analysis, we mainly focused on each pedestrian's speed, distance to other pedestrians, and their extracted emotions: anger, sadness, happiness, and fear.


\section{Results}
\label{sec:results}

This section presents some qualitative analysis of extracted data from individuals of three selected video sequences.

\subsection{Video A}

\begin{figure*}[hbt]
\centering
    \subfigure[china_displacement][Start and end of Video A, highlighting pedestrian 2 (blue), pedestrian 6 (orange) and pedestrian 16 (gray).]{\includegraphics[width=8.0cm]{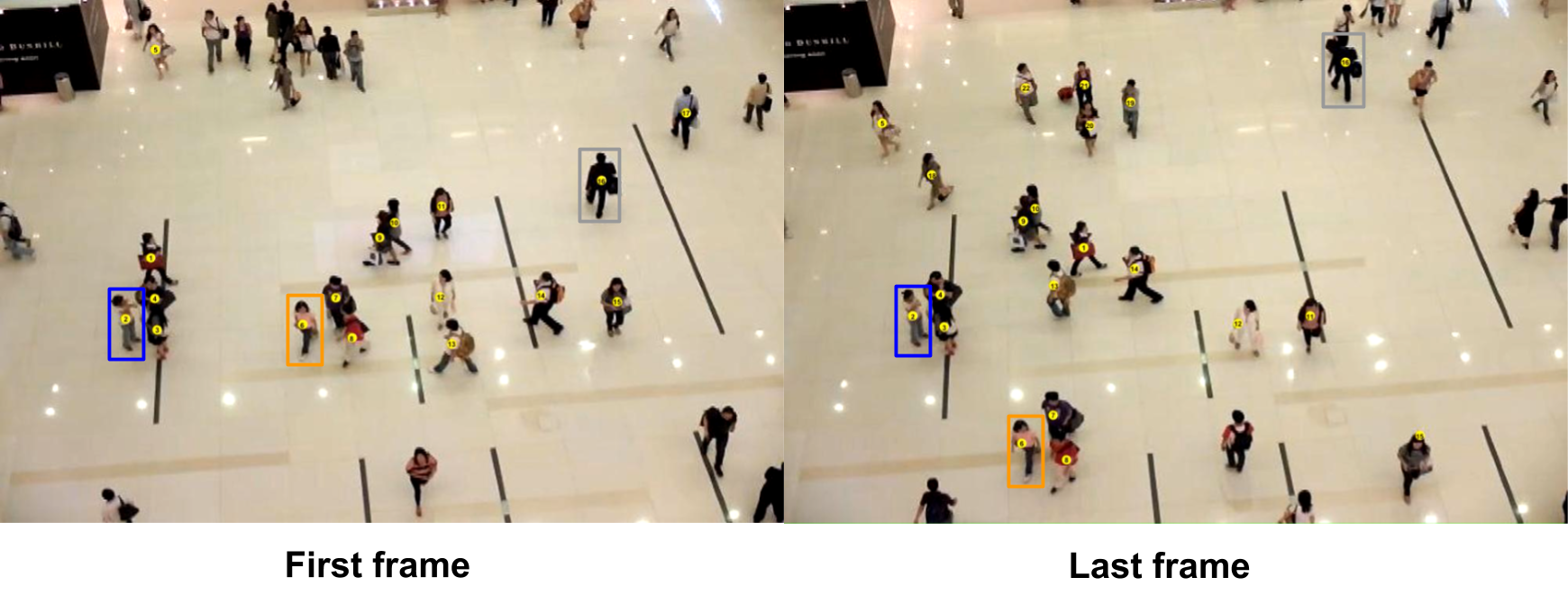}}
    \quad
    \subfigure[china_speed][Speed of pedestrians 2, 6, and 16.]{\includegraphics[width=4.0cm]{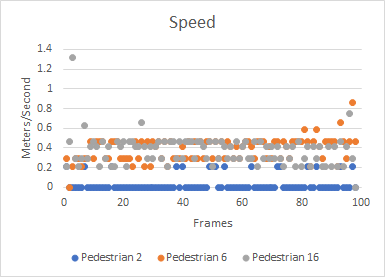}}
    \quad
    \subfigure[china_dist][Average distance to others for pedestrians 2, 6, and 16.]{\includegraphics[width=4.0cm]{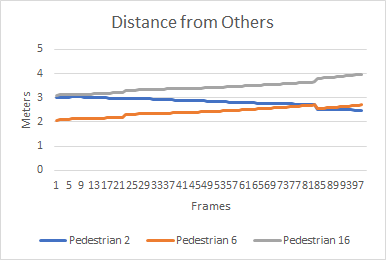}}
    \quad
    \subfigure[china_emo_2][Geometrical emotions for pedestrian 2.]{\includegraphics[width=4.0cm]{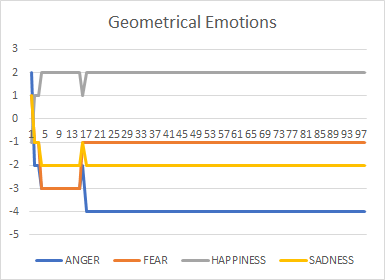}}
\caption{Video A - Usual crowd activity}
\label{fig:china}
\end{figure*}

Video A, depicted in Figure~\ref{fig:china}(a) features footage from a shopping mall in China, taken from the Cultural Crowds database~\cite{cultural_crowds}. It serves as our control, representing usual crowd behavior. There are in total 22 pedestrians in the scene, and no anomaly occurs throughout the duration of the video sequence. We are going to analyze the output of 3 different people, each one displaying a distinct behavior. The first one, pedestrian 2 (blue), standing together with a group of people; the second one, pedestrian 6 (orange), walking in a group; And the third one, pedestrian 16 (gray), walking alone.

We can observe that, in Figure~\ref{fig:china}(b) the speed of each pedestrian stays relatively the same throughout the video. The pedestrians are highlighted in Figure~\ref{fig:china}(a). pedestrian 2's (blue) speed remains close to $0 m/s$ and pedestrians 6's (orange) and 16's (gray) stays around $0.4 m/s$. Figure~\ref{fig:china}(c) shows the average distance of each pedestrian in relation to every other pedestrians in the frame. We can see that although the distance is varying, the change is gradual, suggesting that the crowd is well behaved.

As stated in Section~\ref{sec:method}, Geomind is also able to assign values to certain emotions according to the geometrical information for a given person. In Figure~\ref{fig:china}(d) we can see the progress of these emotions throughout the frames of the video. As expected, no significant change occurs; barring the initial frames, which are not very representative since the behavior up to that point is unstable.

\subsection{Video B}

\begin{figure*}[htb]
\centering
    \subfigure[ACA_displacement][Video B before and during the dispersion event, highlighting pedestrian 1 (blue), pedestrian 2 (orange) and pedestrian 7 (gray).]{\includegraphics[width=8.0cm]{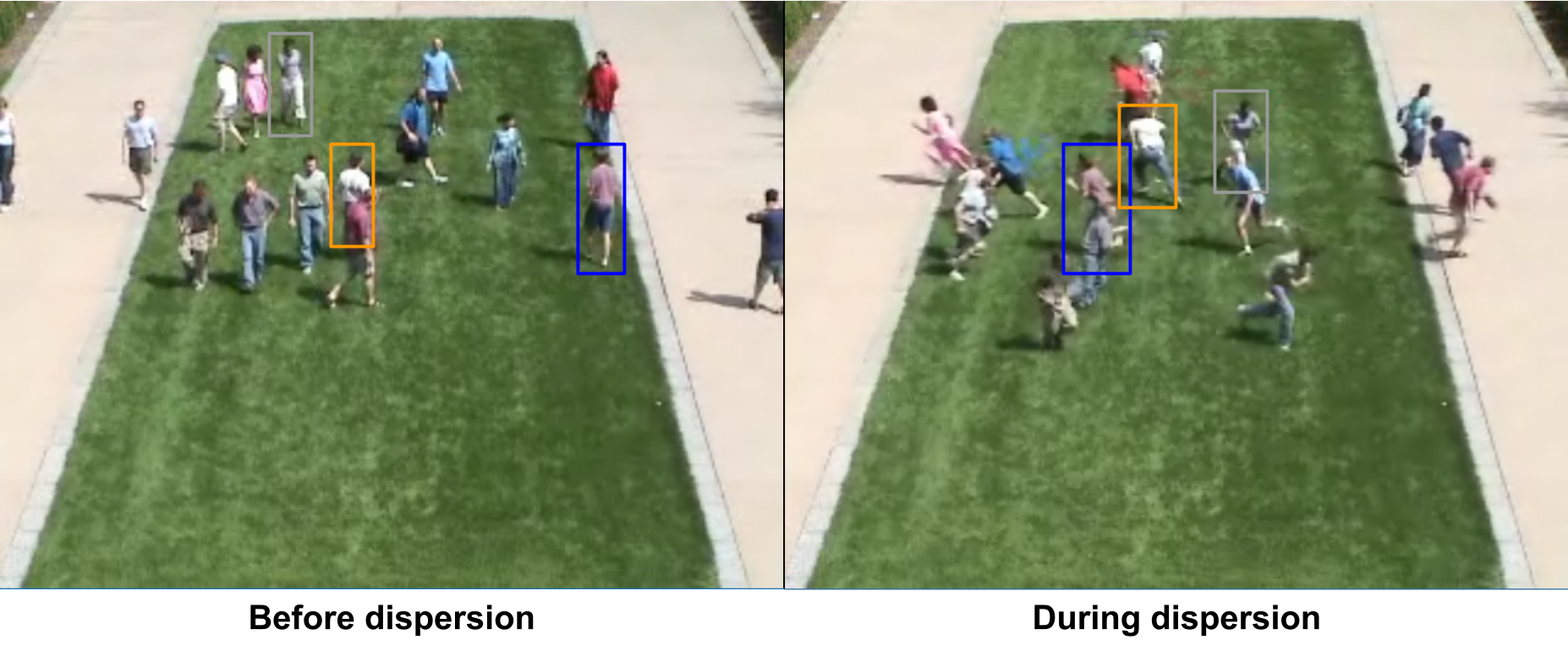}}
    \quad
    \subfigure[ACA_speed][Speed of pedestrians 1, 2, and 7.]{\includegraphics[width=4.0cm]{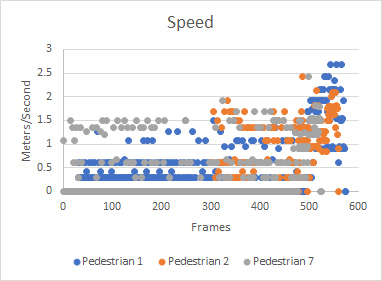}}
    \quad
    \subfigure[ACA_dist][Average distance to others for pedestrians 1, 2, and 7.]{\includegraphics[width=4.0cm]{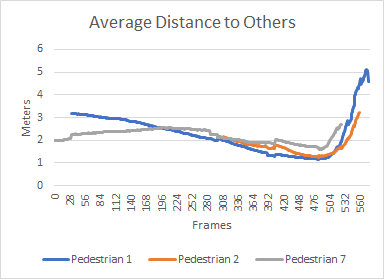}}
    \quad
    \subfigure[ACA_emotion_1][Geometrical emotions for pedestrian 1.]{\includegraphics[width=4.0cm]{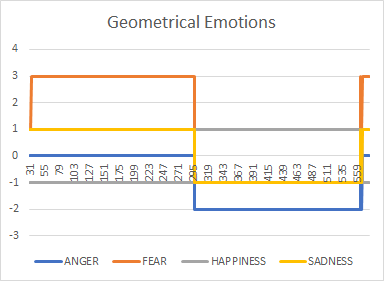}}
    \quad
    \subfigure[ACA_emotion_2][Geometrical emotions for pedestrian 2.]{\includegraphics[width=4.0cm]{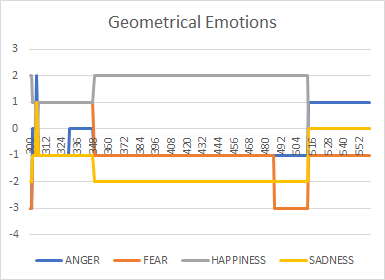}}
\caption{Video B - Crowd dispersion}
\label{fig:unusual_crowd_data}
\end{figure*}

The Detection of Unusual Crowd Activity dataset~\cite{unusual_dataset} features exclusive footage of people acting out unusual crowd behavior; from this dataset, we have selected Video B, depicted in Figure~\ref{fig:unusual_crowd_data}(a). The video starts with each person moving through the scene either alone or in a group; near the end of the scene, all pedestrians suddenly disperse, as if they're fleeing from some incident. We're going to analyze three of the sixteen people present in the scene: pedestrians 1 (blue) and 2 (orange), walking by themselves; in addition to pedestrian 7 (gray), which is walking side-by-side with someone. In Figure~\ref{fig:unusual_crowd_data}(a) we have representative frames from before and after the dispersion. 

The dispersion occurs at frame $470$ and can easily be recognized by observing the speed of the pedestrians, as seen in Figure~\ref{fig:unusual_crowd_data}(b), where the average speed increases by about 1m/s. The event can also be identified from the distance relative to other pedestrians; whilst it was generally changing gradually before the event, it quickly spikes from an average of $1.47 m$ at frame $450$ to about $1.95 m$ in frame $520$.

In Figure~\ref{fig:unusual_crowd_data}(c) and (d) we can observe the emotional data of pedestrians 1 and 2, respectively. In both charts it is possible to see the effects of the event, with fear and sadness rising, whilst happiness decreases; a behavior that corresponds with the situation. 

\subsection{Video C}

\begin{figure*}[htb]
\centering
    \subfigure[fight_displacement][Video B before and after the fight event, highlighting pedestrian 1 (blue), pedestrian 2 (orange), and pedestrian 4 (gray).]{\includegraphics[width=8.0cm]{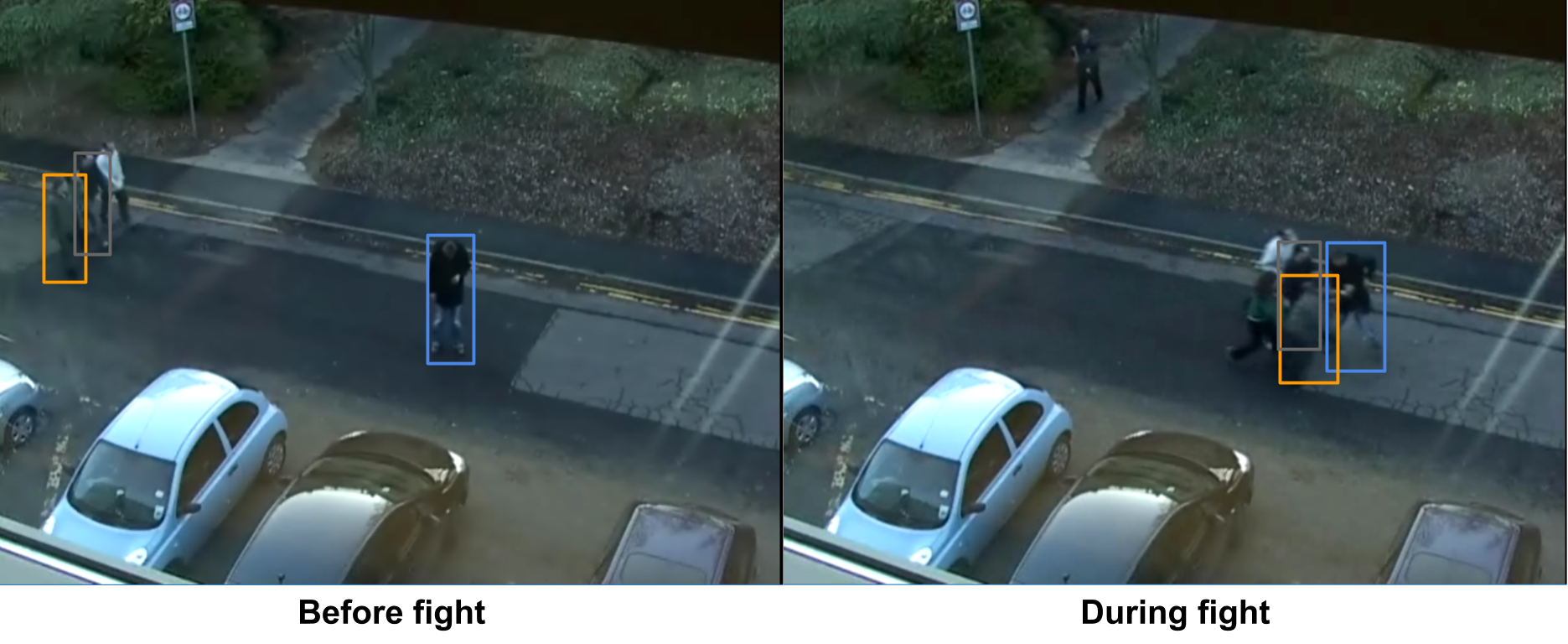}}
    \quad
    \subfigure[fight_speed][Speed of pedestrians 1, 2, and 4.]{\includegraphics[width=4.0cm]{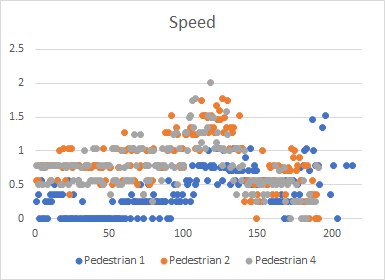}}
    \quad
    \subfigure[fight_dist][Average distance to others for pedestrians 1, 2, and 4.]{\includegraphics[width=4.0cm]{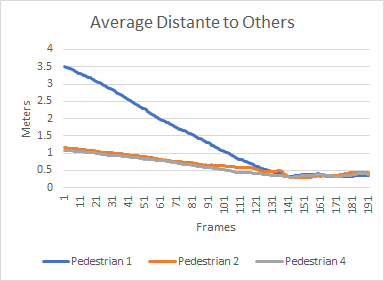}}
    \quad
    \subfigure[fight_emotion_1][Geometrical emotions for pedestrian 1.]{\includegraphics[width=4.0cm]{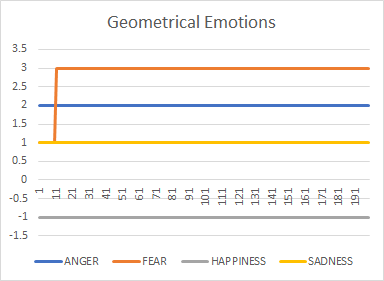}}
    \quad
    \subfigure[fight_emotion_2][Geometrical emotions for pedestrian 2.]{\includegraphics[width=4.0cm]{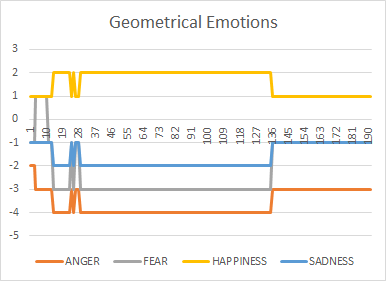}}
\caption{Video C - Fight activity}
\label{fig:fight_data}
\end{figure*}

Video C, portrayed in Figure~\ref{fig:fight_data}(a), belongs to the BEHAVE Interactions Test Case Scenarios dataset~\cite{behave_dataset}, which contains several scenarios of people acting out various interpersonal interactions. We have selected a video of a lone stationary person, pedestrian 1 (blue), who is attacked by four other people, including pedestrians 2 (orange) and 4 (gray); the victim of the attack, then, tries to run away but is eventually tackled to the ground, soon after, the actors leave the frame and thus the video sequence ends. 

In Figure~\ref{fig:fight_data}(c) we see the aggressors approaching pedestrian 1 (blue) through its distance to others, which decreases; curiously, because the aggressors are moving in a group, their average distance doesn't change as drastically. Regarding the pedestrians' velocity, the sudden speed increase as the aggressors approach the victim at the start of the fight can clearly be seen around frame $100$; as the fight progresses and the aggressors take hold of the victim, the speed decreases, as seen in frame $140$; the remaining tumble to the ground features an elevated variation in speed. Although these individual components could not be very telling on their own, their composition could be indicative of a fight or similar event. The combination of slowly decreasing distance to others and a spike in velocity could also signal an attack from a group of people in unison.

Regarding the extracted emotions of the pedestrians in the video, pedestrian 1 emotions, as seen in Figure~\ref{fig:fight_data}(d), remain somewhat constant throughout the sequence. 
Pedestrian 2 exhibits the most favorable results regarding the goal to identify events because, as we can see in Figure~\ref{fig:fight_data}(e), there is a decrease in happiness and an increase in anger and sadness, which fits to the situation at hand. 
The remaining of pedestrians in the video, that is, the other aggressors, don't exhibit any kind of unusual activity.

\section{Final Considerations}
This paper presents an investigation of event detection in video sequences. Our hypothesis is that we can use individual extracted information to detect events without the training phase, once pedestrian data can include personalities and emotions described using GeoMind.
In short, although we've analyzed only a small subset of abnormal events that can happen in a crowd using physical and emotional values, our detection of events changes according to which type of event it is. As this study identified, dispersion events are easier to recognize; whilst fights are much harder, due to the nature of their more nuanced movements. Over the course of this short paper, we have analyzed both normal and abnormal crowd behavior. 
Our results could also be used to validate and feed crowd data for simulated scenarios, such as in digital games, where virtual agents must react realistically to player interaction.
\label{sec:final}

\section*{Acknowledgments}
We would like to thank CAPES, CNPQ, and PUCRS for financially supporting our project.





%


\bibliographystyle{IEEEtran}
\bibliography{example}

\end{document}